\documentclass{ifacconf}

\usepackage{graphicx}      
\usepackage{natbib}        
\usepackage{amsmath,amssymb,amsfonts}
\usepackage{algorithm}

\let\classAND\AND
\let\AND\relax
\usepackage{algorithmic}

\let\AND\classAND
\AtBeginEnvironment{algorithmic}{\let\AND\algoAND}
\usepackage{textcomp}
\usepackage{xcolor}
\usepackage{bm}
\usepackage{pstool}
\usepackage{relsize}
\usepackage{paralist}
\usepackage{siunitx}
\usepackage[inline]{enumitem}
\usepackage{tabularx,booktabs}
\newcolumntype{Y}{>{\centering\arraybackslash}X}
\usepackage{colortbl}
\usepackage{cases}
\usepackage{blindtext}
\usepackage{etoolbox} 

\def\scaspsbi#1#2#3{{#1}\SPSBI{#2}{#3}}

\def\vecspsbi#1#2#3{\vect{#1}\SPSBI{#2}{#3}}
\def\SPSBI#1#2{^{#1}_{#2}}

\def\matr#1{\bm{\uppercase{#1}}}
\def\vect#1{\bm{\lowercase{#1}}}

\def\set#1{\mathcal{\uppercase{#1}}}
\def\tuple#1{#1}

\def\graycel{\cellcolor[gray]{0.8}}

\newcommand{\lst}[1]{\mathsf{#1}}

\newenvironment{lbmatrix}[1]
  {\left[\array{@{}*{#1}{c}@{}}}
  {\endarray\right]}

\newtheorem{defi}{Definition}

\begin{document}
\begin{frontmatter}

\title{Formal Verification of Robotic Contact Tasks via Reachability Analysis\thanksref{footnoteinfo}} 
\thanks[footnoteinfo]{This work has been supported by Siemens AG.}

\author[First]{Chencheng Tang and Matthias Althoff} 

\address[First]{School of Computation, Information and Technology,\\Technical University of Munich, 85748 Garching, Germany \\(e-mails: \{chencheng.tang,althoff\}@tum.de)}

\begin{abstract}                
   Verifying the correct behavior of robots in contact tasks is challenging due to model uncertainties associated with contacts.
   Standard methods for testing often fall short since all (uncountable many) solutions cannot be obtained. 
   Instead, we propose to formally and efficiently verify robot behaviors in contact tasks using reachability analysis, which enables checking all the reachable states against user-provided specifications.
   To this end,
   we extend the state of the art in reachability analysis for hybrid (mixed discrete and continuous) dynamics subject to discrete-time input trajectories.
   In particular, we present a novel and scalable guard intersection approach  
   to reliably compute the complex behavior caused by contacts.
   We model robots subject to contacts as hybrid automata in which crucial time delays are included.
   The usefulness of our approach is demonstrated by verifying safe human-robot interaction in the presence of constrained collisions, which was out of reach for existing methods.
\end{abstract}

\begin{keyword}
Reachability analysis, hybrid systems, guard intersection, formal verification, robotic~contact~tasks, safety.
\end{keyword}

\end{frontmatter}

\section{Introduction} \label{sec_intro}
Robots are often used in contact tasks \citep{ControlRobotsContact2009} where mechanical interactions with other objects have to be considered,
such as deburring, assembly tasks, and applications involving human-robot collaboration \citep{ajoudaniProgressProspectsHuman2018}.
It is essential to guarantee the task specifications for robustifying the process or ensuring safety,
which is commonly fulfilled via extensive and time-consuming testing in industrial practice.
However,
with testing it is impossible to cover all possibilities under uncertainties such as sensor noise, time delays, and mechanical impacts. 
To efficiently provide formal guarantees to contact tasks,
we instead propose to verify robot behaviors using reachability analysis \citep{althoffSetPropagationTechniques2021}, a commonly used method for formal verification.

Reachability analysis computes the set of states that are reachable by a system given its uncertainties.
A task can be formally verified by checking whether the reachable set intersects any unsafe region,
which has been implemented in some contact-task scenarios over the past few years:
\cite{liuOnlineVerificationImpactForceLimiting2021a} verify controllers for human-robot interaction (HRI) during their operation.
However, the approach is not suitable for hard contacts since the system is approximated by continuous-time models.
To precisely describe system behaviors in contact tasks, hybrid (mixed discrete and continuous) dynamics need to be considered due to mechanical impacts and control-mode switching.
With hybrid dynamics considered, \cite{muradoreRoboticSurgery2011} verify the safety of a robotic puncturing task for automatic surgery;
however, the example scenario is restricted to a single transition and a rather small contact force,
while reachability problems are much more challenging in industrial applications, where discrete transition can frequently happen with strong mechanical impacts.
To apply verification results to real contact tasks, another vital factor to consider is the time delay, 
which significantly influences system behaviors \citep{niculescuForceMeasurementTimeDelays1999,surdilovicContactStabilityIssues1996}.
\cite{bresolinOpenProblemsVerification2012} propose a conceptual modeling approach to enclose the effect of time delays.
In contrast to previous work, we consider all the above-mentioned difficulties to formally verify the robot behavior in tasks with significant mechanical impacts by fully considering the inherent hybrid dynamics.

To compute the reachability of a hybrid system, the main challenge is the guard intersection,
i.e., the intersection of reachable sets with guard sets \citep{althoffSetPropagationTechniques2021}.
A popular approach is to use geometric intersection \citep[e.g.,][]{girardZonotopeHyperplaneIntersection2008,althoffComputingReachableSets2010,althoffZonotopeBundlesEfficient2011a}; 
however, for accuracy, these methods usually have to sacrifice scalability with regard to the system dimension.
Some alternatives are proposed to avoid geometric operations:
\cite{althoffAvoidingGeometricIntersection2012a} enclose the intersection by a nonlinear mapping onto the guard.
\cite{bakTimeTriggeredConversionGuards2017} approximate event-based switching by time-triggered switching through dynamics scaling.
However, these approaches have limitations in real-world applications: The former cannot well handle intersections of long duration,
while the latter creates large over-approximations if the reachable set cannot be quickly flattened during the scaling phase.
Furthermore, all the mentioned methods assume a constant input set and cannot consider a user-specified discrete-time input trajectory, which is common in robotics.
Due to these shortcomings, the above-mentioned methods cannot reliably solve reachability problems in the context of contact tasks. 

\begin{figure}[t]
  \centering
  \psfragfig[width=1\columnwidth]{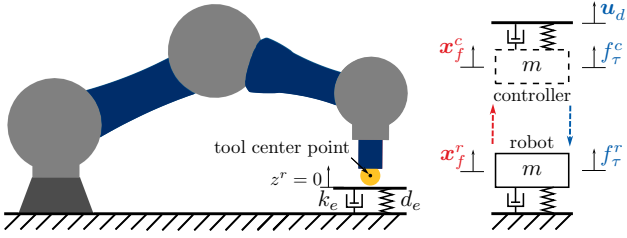}{%
    \psfrag{a}[lc][lc][0.85]{\color[RGB]{0, 92, 171}{$\vecspsbi{u}{}{d}$}}
    \psfrag{b}[lc][lc][0.85]{\color[RGB]{0, 92, 171}{$\scaspsbi{f}{c}{\tau}$}}
    \psfrag{c}[lc][lc][0.85]{\color[RGB]{0, 92, 171}{$\scaspsbi{{f}}{r}{\tau}$}}
    \psfrag{g}[rc][rc][0.7]{\color[RGB]{227, 27, 35}{}}
    \psfrag{n}[rc][rc][0.85]{\color[RGB]{227, 27, 35}{$\vecspsbi{x}{c}{f}$}}
    \psfrag{h}[rc][rc][0.7]{\color[RGB]{227, 27, 35}{}}
    \psfrag{j}[rc][rc][0.85]{\color[RGB]{227, 27, 35}{$\vecspsbi{x}{r}{f}$}}
    \psfrag{0}[cc][cc][0.7]{tool center point}
    \psfrag{r}[cc][cc][0.7]{robot}
    \psfrag{con}[cc][cc][0.7]{controller}
    \psfrag{1}[rc][rc][0.85]{$k_e$}
    \psfrag{2}[lc][lc][0.85]{$d_e$}
    \psfrag{5}[rc][rc][0.85]{$$}
    \psfrag{6}[lc][lc][0.85]{$$}
    \psfrag{m}[cc][cc][0.85]{$m$}
    \psfrag{t}[cc][cc][0.7]{$z^r = 0$}
  }
  \caption{A manipulator collides with a surface.
    The tool center point is located at the center of the sphere and the system reference frame is fixed to the surface.
  }
  \label{fig_scene}
  \vspace{-0.0em}
\end{figure}

This paper presents novel techniques to formally verify the robot behavior in contact tasks.
In contrast to prior work, we consider industrial applications and the above-mentioned challenges associated with them.
The robot dynamics is modeled by hybrid automata with discrete-time input and crucial time delays considered.
Addressing the limitations of existing methods, novel approaches are proposed to reliably compute reachable sets and particularly guard intersections with high accuracy despite good scalability. 
We demonstrate our approach by a typical constrained-contact scenario as shown in Fig.~\ref{fig_scene} \citep{vukobratovicDynamicsContactTasks1999,kirschnerExperimentalAnalysisImpact2021,haddadinRoleRobotMass2008}.

The rest of this paper is structured as follows:
In Sec.~\ref{sec_preandproblem}, we present the research problem after introducing the preliminaries.
Then, we model the example scenario using a hybrid automaton considering time delays in Sec.~\ref{sec_system}.
Sec.~\ref{sec_ra} presents novel methods for the reachability analysis of contact tasks,
which are demonstrated in Sec.~\ref{sec_example} on verifying the collision safety in HRI, in comparison with relevant state-of-the-art approaches.

\section{Preliminaries and Problem Statement} \label{sec_preandproblem}
This section poses the considered problem after recalling the preliminaries on the formal techniques used.

\subsection{Preliminaries}\label{sec_pre}
We denote vectors by bold lowercases (e.g., $\vect{a}$), matrices by bold uppercases (e.g., $\matr{a}$), lists by sans-serif font (e.g., $\lst{A}$), and sets by calligraphic font (e.g., $\set{a}$).
Sets in this work are represented mainly by zonotopes \citep{kuhnRigorouslyComputedOrbits1998}:
\begin{defi}[Zonotope]\label{def_zonotope}
  Given a center $\vect{c} \in \mathbb{R}^n$ and a generator matrix $\matr{G}=[\vect{g}_{1},\ldots,\vect{g}_{p}] \in \mathbb{R}^{n\times p}$,
  a zonotope is defined as
  \begin{equation*}
        \set{Z} = 
        (\vect{c},\matr{G}) :=       \left\{ \vect{x} = \vect{c} + \sum_{i = 1}^{p} \beta_i\,\vect{g}_{i} \,\middle|\, \beta_i\in[-1,1] \right\}.
  \end{equation*}
\end{defi}
On zonotopes, many operations can be exactly and efficiently computed \citep{althoffSetPropagationTechniques2021},
such as Minkowski addition \mbox{$\set{A}\oplus\set{B} := \left\{\vect{a}+\vect{b} \,\middle|\, \vect{a}\in\set{A},\vect{b}\in\set{B}\right\}$}
and linear transformation $\matr{M}\otimes\set{A} := \left\{\matr{M}\vect{a} \,\middle|\, \vect{a}\in\set{A}\right\}$.
To evaluate the volume of a zonotope, we use the $n^{th}$ root of its interval-hull volume \citep{kuhnRigorouslyComputedOrbits1998}.

Contact tasks are formalized as hybrid automata, with a definition similar to \cite{kochdumperReachabilityAnalysisHybrid2020}:
\begin{defi}[Hybrid Automaton]\label{def_ha}
  A hybrid automaton $\tuple{H}$ is defined by $m$ discrete states referred to as locations $(\tuple{L}_1,\dotsc,\tuple{L}_m)$, where each location $\tuple{L}_i$ consists of:
  \begin{itemize}
    \item A flow function $\dot{\vect{x}}=f_i(\vect{x},\vect{u})$ describing the continuous dynamics in the location, with continuous state $\vect{x}$ and input $\vect{u}$.
    \item An invariant set $\set{S}_i\subset\mathbb{R}^n$ describing the region where the flow function is valid.
    \item A list of transitions $\lst{T}_i=(T_1,\dotsc, T_q)$. Each transition $T_j=(\set{G}_j, r_j(\vect{x},\vect{u}),s_i,d_i)$ contains a guard set \mbox{$\set{G}_j\subset\mathbb{R}^n$}, a jump function $r_j:\mathbb{R}^n\rightarrow\mathbb{R}^n$, and $s_j, d_j$ which are indices of the source and target location, respectively.
  \end{itemize}
\end{defi}

\begin{figure}[t]
  \vspace{0em}
  \centering
  \psfragfig[width=0.9\columnwidth]{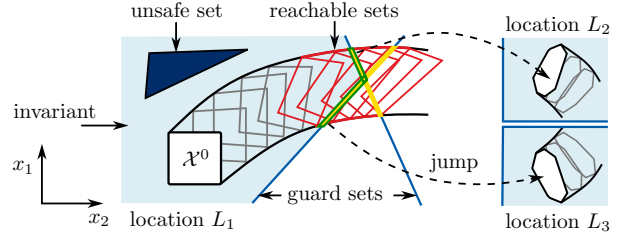}{%
    \psfrag{a}[cc][cc][0.8]{unsafe set}
    \psfrag{b}[cc][cc][0.8]{reachable sets}
    \psfrag{c}[cc][cc][0.8]{jump}
    \psfrag{d}[cc][cc][0.8]{$\set{X}^0$}
    \psfrag{e}[lc][cc][0.8]{guard intersection}
    \psfrag{g}[rc][cc][0.8]{invariant}
    \psfrag{h}[cc][cc][0.8]{location $\tuple{L}_1$}
    \psfrag{m}[cc][cc][0.8]{location $\tuple{L}_2$}
    \psfrag{n}[cc][cc][0.8]{location $\tuple{L}_3$}
    \psfrag{j}[lc][cc][0.8]{guard sets}
    \psfrag{1}[cc][cc][0.8]{$x_1$}
    \psfrag{2}[lc][cc][0.8]{$x_2$}
  }
  \vspace{0em}
  \caption{State evolution in a hybrid automaton.
  }
  \vspace{0em}
  \label{fig_hasemantics}
\end{figure}

We consider urgent transition semantics \citep{henzingerHyTechCornellHybrid1995} and define guards as hyperplanes.
The evolution of a hybrid automaton is illustrated in Fig.~\ref{fig_hasemantics} and is informally described as follows:
Starting from an initial location $\tuple{L}^0$ with initial continuous state $\vect{x}^0 \in \set{X}^0$, $\vect{x}(t)$ evolves following the flow function of $\tuple{L}^0$.
When $\vect{x}(t)$ hits the guard set of a transition, the state transits to the corresponding target location in zero time.
In case several guard sets are hit at the same time, the transition is chosen non-deterministically.
After a transition, the continuous state is updated by the jump function and the result becomes the initial state in the new location where the evolution continues.
We denote the trajectory of the continuous state for the evolution described above by $\xi(t,\tuple{L}^0,\vect{x}^0,\vect{u}(\cdot))$, 
where $\vect{u}(\cdot)$ is an input trajectory.

We are interested in the continuous reachable set of hybrid automata \citep{kochdumperReachabilityAnalysisHybrid2020}:
\begin{defi}[Reachable Set]
  Given an initial location $\tuple{L}^0$, a set of initial continuous states $\set{X}^0$,
  and an input set $\set{U}$,
  the continuous reachable set $R^e(t)$ of a hybrid automaton at time $t$ is:
  \begin{align*}
    \set{R}^e(t)=\left\{\xi(t,\tuple{L}^0,\vect{x}^0,\vect{u}(\cdot)) \,\middle|\, \right. & \vect{x}^0 \in \set{X}^0, \\
    &\left. \forall \tau \in [0, t] : \vect{u}(\tau) \in \set{u} \vphantom{\xi(t,\tuple{L}^0,\vect{x}^0,\vect{u})}\right\}
  \end{align*}
  while for a time interval $[t_1,t_2]$:
  \begin{equation*}
    \mathcal{R}^e([t_1,t_2]) = \bigcup_{t \in [t_1,t_2]}\mathcal{R}^e(t).
  \end{equation*}
\end{defi}
Since the exact reachable set $\set{R}^e(t)$ can only be computed for a limited class of hybrid automata \citep{lafferriereNewClassDecidable1999},
we compute tight over-approximations \mbox{$\set{R}(t)\supset\set{R}^e(t)$}.
The computation for all times $[0,t_f]$ is simply the union of the reachable sets of subsequent time intervals.
Accordingly, we compute \mbox{$\set{R}_{0,1},\ldots,\set{R}_{N-1,N}, N\in\mathbb{N}$} such that $\set{R}_{k,k+1}\supset\set{R}^e([k\Delta\tau,(k+1)\Delta\tau])$,
where the time step size $\Delta\tau=t_f/N$ is a tunable parameter largely influencing the computation accuracy \citep{wetzlingerAdaptiveParameterTuning2020}.
The basic procedure of computation is shown in Fig.~\ref{fig_basicproc} in black boxes,
which repeats in each location:
\begin{enumerate}
  \item Starting from the initial set, compute the reachable sets of subsequent time intervals $\set{R}_{k,k+1}$ until the reachable set $\set{R}_k$ of time point $k\Delta\tau$ leaves the invariant or the time horizon $t_{f}$ is reached.

  \item When the reachable sets intersect a guard set $\set{G}_j$ from time $t_{h}$ to $t_{l}$ (bounded in this work),
  compute the intersection $\set{I}_j$ of $\set{R}([t_{h}, t_{l}])$ with $\set{G}_j$.
  When multiple guard sets (e.g., $\set{G}_1,\set{G}_2$) are hit by the reachable sets, separately compute $\set{I}_1$ and $\set{I}_2$, as illustrated by the yellow sets in Fig.~\ref{fig_hasemantics}.
  In that case, the intersection $\set{I}_j$ can be tightened by removing the parts which trigger other transitions before intersecting $\set{G}_j$, resulting in the green sets in Fig.~\ref{fig_hasemantics}.
  In this work, the pruning can be realized by intersecting $\set{I}_j$ with the invariant.  
  \item For each intersected guard, propagate the intersection using the jump function, the result of which becomes the initial set in the target location.
\end{enumerate}

The input in a robot task commonly contains a user-specified discrete-time trajectory $\lst{U}_d=(\vect{u}_{d,0}, \ldots, \vect{u}_{d,f})$, where $f$ denotes the final step.
Given a trajectory sampled at time points $t_k = kT, k\in\mathbb{N}_0$ with time increment $T \in \mathbb{R}_{>0}$,
the continuous-time input is realized by zero-order hold as
\begin{equation} \label{eq_ud}
\forall t \in [t_k, t_{k+1}) : \vect{u}_{d}(t) = \vect{u}_{d,k}.
\end{equation}
The above input requires that the propagation of reachable sets is synchronized with time---this has not yet been realized for hybrid systems so far,
because time synchronization is not maintained after guard intersections as will be detailed later.


\subsection{Problem Statement} \label{sec_pro}
We verify the robot behavior in contact tasks modeled as hybrid automata.
Given an initial location $\tuple{L}^0$, an initial set $\set{X}^0$, an input set $\set{U} = f^u(\vect{u}_d(t),\set{U}_u)$ with a discrete-time input trajectory $\vect{u}_d$ as well as an uncertainty set $\set{U}_u$,
and an unsafe set $\set{F}$ representing the violation of task specifications,
the verification problem for a time interval $[0,t_{f}]$ is:
\begin{gather*}
  \mathrm{safe} \Leftrightarrow \nexists \mkern4mu \xi(t,\tuple{L}^0,\vect{x}^0,\vect{u}(\cdot)) \in \set{F} \mkern185mu \\
  \mkern52mu \text{with} \mkern4mu t \in [0,t_{f}], \vect{x}(0) \in \set{X}^0, \vect{u}(t) \in f^u(\vect{u}_d(t),\set{U}_u).
\end{gather*}
By using reachability, we can verify safety as follows:
\begin{equation*}
  \set{R}([0,t_f]) \cap \set{F} = \emptyset \Rightarrow \mathrm{safe} .
\end{equation*}

\section{Modeling robot dynamics With Delays} \label{sec_system}
We model a typical scenario of constrained contact as shown in Fig.~\ref{fig_scene},
where a robot under compliant control hits a surface.

\subsection{Robot and Controller} \label{mss}
A robotic system typically consists of the robot itself and its controller.
We denote the variables associated with the robot by superscript $r$
and the variables associated with the controller by superscript $c$.
For the system in Fig.~\ref{fig_scene}, the controller uses 
$\vect{x}_f^c =
  \setlength{\arraycolsep}{2.5pt}
  \begin{bmatrix}
    {\vecspsbi{z}{c}{}}       &
    {\vecspsbi{\dot{z}}{c}{}}
  \end{bmatrix}$
which is the delayed measurement of the system states
$\vect{x}_f^r =
  \setlength{\arraycolsep}{2.5pt}
  \begin{bmatrix}
    {\vecspsbi{z}{r}{}}       &
    {\vecspsbi{\dot{z}}{r}{}}
  \end{bmatrix}$, 
where $\vecspsbi{z}{r}{} \in \mathbb{R}^{6}$ is the position of the tool center point (TCP).
The user-specified input trajectory is
$\vect{u}_d =
  \setlength{\arraycolsep}{2.5pt}
  \begin{bmatrix}
    \vect{z}_d       &
    \dot{\vect{z}}_d &
    \ddot{\vect{z}}_d
  \end{bmatrix}$, where
$\vect{z}_d,
  \dot{\vect{z}}_d,
  \ddot{\vect{z}}_d \in \mathbb{R}^{6}$
are respectively the desired position, velocity, and acceleration for the TCP.

The robot dynamics in Cartesian coordinates is the result of actuation force $\vect{f}_\tau^r$ and contact force $\vect{f}_e^r$
applied to the effective mass $\matr{\Lambda}(\vect{q})$ \citep[eq. 6]{khatibInertialPropertiesRobotic1995}, which depends on the joint position vector $\vect{q}$:
  \begin{equation} \label{eq_law_sl_o}
    \scaspsbi{\ddot{\vect{z}}}{r}{} = \matr{\Lambda}(\vect{q})^{-1}({\vect{f}_\tau^r + \vect{f}_e^r}).
  \end{equation}
We only consider nonsingular configurations of robots and the translational part of the dynamics, since our main interest is in the normal force.
Assuming that the posture of robots barely changes during contact,
the effective mass of the robot along one direction can be approximated by a constant value $m$ \citep{khatibInertialPropertiesRobotic1995}.
Accordingly, the three Cartesian dimensions can be decoupled, leading to a scalar representation of the dynamics in the normal-force direction: 
  \begin{equation} \label{eq_law_sl}
    \scaspsbi{\ddot{{z}}}{r}{} = m^{-1}({{f}_\tau^r + {f}_e^r}).
  \end{equation}

Two typical modes are considered for the robot controller: 
One mode is a Cartesian impedance controller \citep[eq. 19]{albu-schafferCartesianImpedanceControl2003},
with desired stiffness and damping factors $k_t, d_t$:
\begin{equation} \label{eq_law_ma}
  \scaspsbi{{f}}{c}{\tau} = m\ddot{{z}}_d+d_t(\dot{{z}}_d-\dot{{z}}^c)+k_t({{z}}_d-{{z}}^c).
\end{equation}
The other mode is a collision reaction mode \citep{delucaIntegratedControlPHRI2012} with damping factor $d_r$, which is triggered once the controller receives a contact force ${f}^c_e$ exceeding the user-defined threshold ${f}_t$:
\begin{equation} \label{eq_collisionreaction}
  \scaspsbi{{f}}{c}{\tau} = -d_r \scaspsbi{\dot{{z}}}{c}{}.
\end{equation}

To model the contact force ${f}_e^r$, we adopt the widely used Kelvin-Voigt contact model \citep{achhammerImprovementModelmediatedTeleoperation2010},
as it provides reasonable results for most cases despite its simplicity \citep{gilardiLiteratureSurveyContact2002}.
Given $l$ as the radius of the sphere in Fig.~\ref{fig_scene}, $k_e, d_e$ respectively as the stiffness and damping coefficients of the contact,
we have:
\begin{subequations} \label{eq_fe}
  \begin{numcases}{f_e^r =}
    0 & if $z^r \geq l$ \label{eq_fe_nc} \\
    -k_e (z^r - l) - d_e \dot{z}^r & otherwise. \label{eq_fe_wc}
  \end{numcases}
\end{subequations}

With \eqref{eq_law_sl}-\eqref{eq_fe}, we obtain the system dynamics in the direction of the normal force.
Please note that the one-dimensional abstraction is commonly used in relevant studies \citep[e.g.,][]{muradoreRoboticSurgery2011,liuOnlineVerificationImpactForceLimiting2021a,achhammerImprovementModelmediatedTeleoperation2010}.
Despite its simplicity, it is proven sufficient in describing higher-dimensional cases when the nonlinear dynamics are decoupled and compensated for \citep{niculescuForceMeasurementTimeDelays1999}.

\subsection{Delay Formulation} \label{sec_delayfor}
We consider two crucial time delays in the system:
a robot-to-controller delay $d_{rc}$ and a controller-to-robot delay $d_{cr}$.
The input is only affected by $d_{cr}$, while the state is affected by both delays.
Accordingly, the actual actuation force at time $t$ is:
\begin{equation}\label{eq_delay_controlsignal}
  \begin{aligned}
    f_\tau^r(t) & =
    f_\tau^c(\vect{u}_d(t-d_{cr}), \vect{x}_f^r (t-d_{cr}-d_{rc}))                                 \\
                & = f_\tau^c(\vect{u}_d(t-d_1), \vect{x}_f^r (t-d_2)) = f_\tau^c(\hat{\vect{u}}_d(t), \hat{\vect{x}}_f^r(t)),
  \end{aligned}
\end{equation}
where the delays are rewritten as $d_1, d_2$, and the delayed variables are denoted by a hat.

While the delayed input $\hat{\vect{u}}_{d}$ can be easily obtained with a predefined input trajectory,
we approximate the delay differential equation of states $\dot{\hat{\vect{x}}}_f^r(t) = \dot{{\vect{x}}}_f^r(t-d_2)$ using the Padé approximation \citep{molerNineteenDubiousWays2003}
to employ powerful methods for the reachability analysis of ordinary differential equations.
Because the time delay is rather short in modern robotic systems and to avoid numerous additional state variables (reduces the computation time of reachability analysis),
we employ a Padé approximation of first-order:
\begin{equation} \label{eq_delay_app2}
  \dot{\hat{\vect{x}}}_f^r(t) \cong - \frac{2{\hat{\vect{x}}}_f^r(t)}{d_2} + \frac{2{{\vect{x}}}_f^r(t)}{d_2}  - \dot{{\vect{x}}}_f^r(t).
\end{equation}

\subsection{Overall Model} \label{cs}
\begin{figure}[h]
  \centering
  \psfragfig[width=1\columnwidth]{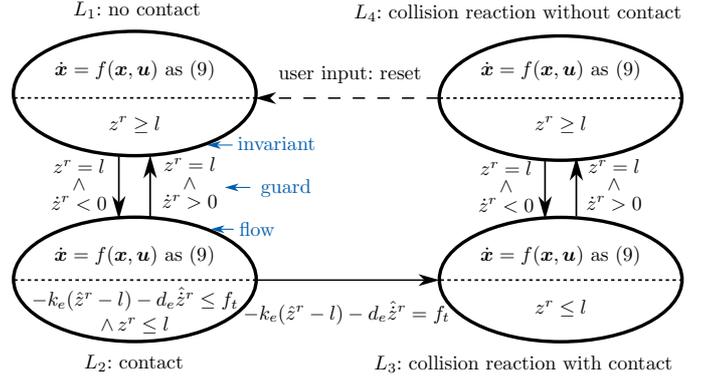} {%
    \psfrag{1}[cc][cc][0.75]{\shortstack[c]{$\dot{\vect{x}}=f(\vect{x},\vect{u})$ as \eqref{eq_ss_all}}}
    \psfrag{2}[cc][cc][0.75]{$ z^r \geq l $}
    \psfrag{3}[cc][cc][0.75]{\shortstack[c]{$\dot{\vect{x}}=f(\vect{x},\vect{u})$ as \eqref{eq_ss_all}}}
    \psfrag{4}[cc][cc][0.75]{\shortstack[c]{$-k_e (\hat{z}^r-l) - d_e \hat{\dot{z}}^r \leq f_t $ \\ $\wedge \mkern3mu z^r \leq l$}}
    \psfrag{5}[cc][cc][0.75]{\shortstack[c]{$\dot{\vect{x}}=f(\vect{x},\vect{u})$ as \eqref{eq_ss_all}}}
    \psfrag{6}[cc][cc][0.75]{$ z^r \leq l $}
    \psfrag{7}[cc][cc][0.75]{\shortstack[c]{$\dot{\vect{x}}=f(\vect{x},\vect{u})$ as \eqref{eq_ss_all}}}
    \psfrag{8}[cc][cc][0.75]{$ z^r \geq l $}
    \psfrag{a}[rt][cc][0.75]{\shortstack[c]{$z^r = l$ \\ $\wedge$ \\ $\dot{z}^r<0$}}
    \psfrag{b}[lt][cc][0.75]{\shortstack[c]{$z^r = l$ \\ $\wedge$ \\ $\dot{z}^r>0$}}
    \psfrag{c}[ct][cc][0.75]{$-k_e (\hat{z}^r-l) - d_e \hat{\dot{z}}^r = f_t$}
    \psfrag{y}[cb][cc][0.75]{\shortstack[c]{user input: reset}}
    \psfrag{d}[rt][cc][0.75]{\shortstack[c]{$z^r = l$ \\ $\wedge$ \\ $\dot{z}^r<0$}}
    \psfrag{e}[lt][cc][0.75]{\shortstack[c]{$z^r = l$ \\ $\wedge$ \\ $\dot{z}^r>0$}}
    \psfrag{f}[cb][cc][0.75]{$\tuple{L}_1$: no contact}
    \psfrag{g}[cb][cc][0.75]{$\tuple{L}_4$: collision reaction without contact}
    \psfrag{h}[ct][cc][0.75]{$\tuple{L}_2$: contact}
    \psfrag{i}[ct][cc][0.75]{$\tuple{L}_3$: collision reaction with contact}
    \psfrag{u}[lc][cc][0.75]{\color[RGB]{0, 92, 171}{flow}}
    \psfrag{v}[lc][cc][0.75]{\color[RGB]{0, 92, 171}{invariant}}
    \psfrag{w}[lc][cc][0.75]{\color[RGB]{0, 92, 171}{guard}}
    \psfrag{x}[lc][cc][0.75]{\color[RGB]{0, 92, 171}{jump}}
  }
  \caption{Hybrid automaton of the system.}
  \vspace{0em}
  \label{fig_ha}
\end{figure}

We model the overall system using a hybrid automaton as shown in Fig.~\ref{fig_ha}.
The discrete state transits to $\tuple{L}_2$ from $\tuple{L}_1$ if the robot approaches the surface
and transits back to $\tuple{L}_1$ when the robot leaves the surface.
If the delayed contact force $\hat{f}_e^r(t) = -k_e \hat{z}^r - d_e \hat{\dot{z}}^r$ breaches the threshold $f_t$, the collision reaction is triggered (transition from $\tuple{L}_2$ to $\tuple{L}_3$) to decrease the potential damage. 
The control mode stays in collision reaction since then,
and the transitions between $\tuple{L}_3$ and $\tuple{L}_4$ are similar to those between $\tuple{L}_1$ and $\tuple{L}_2$.
The task can be reset by user input, as illustrated by the dotted arrow in Fig.~\ref{fig_ha}.

The state-space model of the system is defined given the state vector
$ \vect{x} =
  \begin{bmatrix}
    \vecspsbi{x}{r}{f}       &
    \vecspsbi{\hat{x}}{r}{f} &
    t
  \end{bmatrix}^\mathsf{T}
  =
  \begin{bmatrix}
    z^r       &
    \dot{z}^r &
    \hat{z}^r       &
    \hat{\dot{z}}^r &
    t
  \end{bmatrix}^\mathsf{T}
$,
and the input vector
$ \vect{u} = \hat{\vect{u}}_{d} {}^\mathsf{T} = 
  \setlength{\arraycolsep}{2.5pt}
  \begin{bmatrix}
    \hat{z}_d       &
    \hat{\dot{z}}_d &
    \hat{\ddot{z}}_d
  \end{bmatrix}^\mathsf{T}
$.
The state variable $t$ serves as a clock. Having a clock associated to the state allows us to track the time by projecting a state vector to time (operator $\text{\tt{proj}}_t$):
\begin{equation*}
  t = \text{\tt{proj}}_t(\vect{x}),
\end{equation*}
which will be used in Sec.~\ref{sec_disctraj} to address the time synchronization problem mentioned in Sec.~\ref{sec_pre}.
Accordingly, the flow function is obtained as
\begin{equation} \label{eq_ss_all}
      \dot{\vect{x}} =
      \setlength{\arraycolsep}{2.5pt}
      \begin{lbmatrix}{4}
        \matr{A}_1   & \matr{A}_2 & \normalsize{\matr{0}}\\[3.5pt]
        \frac{2}{d_2}\matr{I}_2-\matr{A}_1 & -\frac{2}{d_2}\matr{I}_2-\matr{A}_2  & \normalsize{\matr{0}}\\[3pt]
        \normalsize{\matr{0}} & \normalsize{\matr{0}} & \normalsize{\matr{0}}
      \end{lbmatrix}
      \normalsize\vect{x}
      +
      \setlength{\arraycolsep}{2pt}
      \begin{lbmatrix}{1}
        \normalsize{\matr{0}} \\[5pt]
        \normalsize{\matr{0}} \\[3.5pt]
        1
      \end{lbmatrix}
      +
      \setlength{\arraycolsep}{2pt}
      \begin{lbmatrix}{1}
        \matr{B}_1 \\[5pt]
        -\matr{B}_1 \\[3.5pt]
        \normalsize{\matr{0}}
      \end{lbmatrix}
      \normalsize\vect{u}
    .
  \end{equation}
The first line, i.e.,
$\dot{\vect{x}}_{f}^r = \begin{lbmatrix}{3}
  \matr{A}_1   & \matr{A}_2
\end{lbmatrix}
\,
\begin{lbmatrix}{3}
  \vecspsbi{x}{r}{f} & \vecspsbi{\hat{x}}{r}{f}
\end{lbmatrix}^\mathsf{T} +  \matr{B}_1\vect{\hat{u}}_{d} {}^\mathsf{T}
$,
represents the robot dynamics,
where the values in the matrices vary for each location and can be easily derived from \eqref{eq_law_sl}-\eqref{eq_delay_controlsignal}.
The second line is simply the delay formulation \eqref{eq_delay_app2},
and the third line models the clock.

\section{Reachability Analysis} \label{sec_ra}

We introduce a new procedure for reachability analysis with user-specified discrete-time input trajectories.
Then, we present novel approaches for reliably computing guard intersections with high accuracy despite good scalability.

\subsection{Extension for Discrete-Time Input Trajectories} \label{sec_disctraj}
In Sec.~\ref{sec_pre}, we explained the basic procedure for the reachability analysis of hybrid systems with the input set $\set{U}$ given,
and proposed \eqref{eq_ud} to extract the input from a discrete-time trajectory based on time $t$.
However, set propagations are not synchronized with the discretized time $k\Delta\tau$ after guard intersection,
which unifies the sets of states that hit the guard at different times and thus introduces time uncertainty.
Accordingly, we associate a clock with the state, as illustrated in Sec.~\ref{cs},
and introduce two additional steps below for reachability analysis with discrete-time input trajectories. 

\begin{figure}[h]
  \centering
  \psfragfig[width=1\columnwidth]{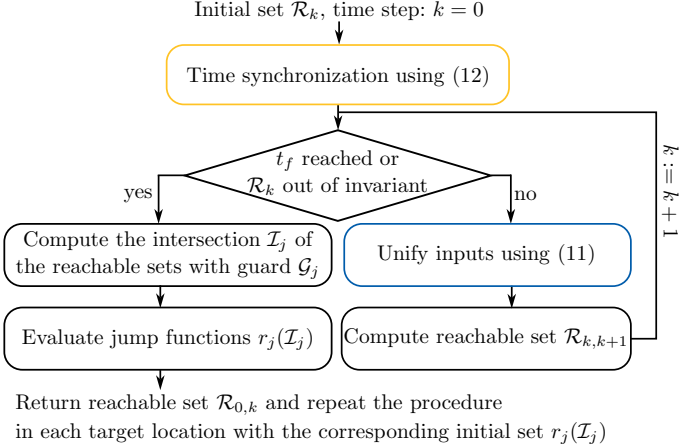}{%
    \psfrag{0}[cc][cc][0.8]{Initial set $\set{R}_k$, time step: $k = 0$}
    \psfrag{a}[cc][cc][0.8]{Time synchronization using \eqref{eq_timehyperplane}}
    \psfrag{2}[cc][cc][0.8]{\shortstack[c]{Unify inputs using \eqref{eq_u}}}
    \psfrag{3}[cc][cc][0.8]{\shortstack[c]{Compute the intersection $\set{I}_j$ of\\the reachable sets with guard $\set{G}_j$}}
    \psfrag{4}[cc][cc][0.8]{Evaluate jump functions $r_j(\set{I}_j)$}
    \psfrag{y}[cc][cc][0.8]{yes}
    \psfrag{n}[cc][cc][0.8]{no}
    \psfrag{5}[cc][cc][0.8]{\shortstack[c]{Return reachable set $\set{R}_{0,k} \cap \set{S}$\\ and $r(\set{I}_j)$ as the initial set for location $\tuple{L}_j$}}
    \psfrag{6}[lc][cc][0.8]{$k:=k+1$}
    \psfrag{7}[cc][cc][0.8]{Compute reachable set $\set{R}_{k,k+1}$}
    \psfrag{8}[cc][cc][0.8]{\shortstack[c]{$t_{f}$ reached or \\ $\set{R}_k$ out of invariant}}
    \psfrag{9}[lc][lc][0.8]{\shortstack[l]{Return reachable set $\set{R}_{0,k}$ and repeat the procedure \\ in each target location with the corresponding initial set $r_j(\set{I}_j)$}}
  }
  \caption{Procedure for reachability analysis with discrete-time input trajectories.}
  \vspace{-0em}
  \label{fig_basicproc}
\end{figure}

To precisely enclose the possible inputs for each iteration, we introduce the step in the blue box in Fig.~\ref{fig_basicproc}.
At the beginning of iteration $k$, we compute the time interval associated with $\set{R}_{k,k+1}$ to capture the time uncertainty introduced by guard intersection:
\begin{equation}
  \set{T} = [\mathrm{inf}(\text{\tt{proj}}_t(\set{R}_k)),\mathrm{sup}(\text{\tt{proj}}_t(\set{R}_k))+\Delta\tau].
\end{equation}
We enclose the discrete-time inputs during the time interval $\set{T}$ by a zonotope.
For the system modeled in Sec.~\ref{sec_system}, the enclosure is computed as
\begin{equation} \label{eq_u}
  \set{U}=\set{U}_u \oplus \left\{\vect{u}_d(t-d_1)  \,\middle|\,  t \in \set{T}\right\},
\end{equation}
where the input uncertainty $\set{U}_u \subset \mathbb{R}^3$ is represented by a zonotope.

Obviously, the time uncertainty enlarges the input set $\set{U}$ in \eqref{eq_u} and causes unnecessary over-approximation.
To address this, we remove the time uncertainty using the step in the yellow box in Fig.~\ref{fig_basicproc}.
When the state enters a location after intersection (intersection set $\set{I}$), 
we compute the start and the end of the intersection period $t_{h},t_{l}$ as the infimum and supremum of $\text{\tt{proj}}_t(\set{I})$.
Then, we compute the set propagations for a period of $t_{l}-t_{h}$ with a time step size $\Delta\tau=(t_{l}-t_{h})/N$ and obtain the reachable set \mbox{$\set{R}_{0,N}$}.
The set $\set{R}_{0,N}$ contains all the states reachable at $t_{l}$ in the current location,
and thus we can compute $\set{R}(t_{l})$ as
\begin{equation} \label{eq_timehyperplane}
  \set{R}(t_{l}) = \set{R}_{0,N} \cap \left\{\vect{x}  \,\middle|\,  \text{\tt{proj}}_t(\vect{x}) = t_{l}\right\}.
\end{equation}
By taking $\set{R}(t_{l})$ as the initial set, the subsequent set propagations are time synchronized.
We apply the introduced technique selectively by checking the magnitude of $t_{l}-t_{h}$ a priori.

\subsection{Guard Intersection} \label{gi}
The idea of our novel guard-intersection approach is intuitive: we select and combine complementary approaches that are scalable with the system dimension.
Three approaches are adopted:
hyperplane mapping \citep{althoffAvoidingGeometricIntersection2012a}, dynamics scaling \citep{bakTimeTriggeredConversionGuards2017},
and an efficient geometric approach implemented in CORA \citep{althoffIntroductionCORA20152015}.

We first recall the three relevant approaches.
The geometric approach represents all the sets intersecting the guard $\set{P}_1,\ldots,\set{P}_p$ by constrained zonotopes \citep{scottConstrainedZonotopesNew2016},
which are closed under intersection with hyperplanes.
Then the intersections between $\set{P}_i$ and the guard set $\set{G}$ are enclosed by their interval hulls for efficiently computing their union:
\begin{equation*}
  \set{I}_g := \bigcup_{i = 1}^{p}  \text{\tt{intervalHull}}(\set{P}_i \cap \set{G}).
\end{equation*}
The operator $\text{\tt{intervalHull}}$ computes the tightest interval enclosure of a constrained zonotope by solving $2n$ linear programs,
which determines the computational complexity of the geometric approach.
A trade-off between accuracy and scalability usually has to be found for geometric approaches.
The above method is designed to sacrifice accuracy for efficiency;
due to the strong dependencies between states in contact dynamics, accurate geometric intersection would be costly.

In contrast to geometric intersection, the mapping method can tightly enclose guard intersection despite good scalability with complexity $\mathcal{O}(n^6)$ \citep[p.~2]{althoffAvoidingGeometricIntersection2012a}.
The mapping method abstracts a system to a state-dependent constant flow $\dot{\vect{x}} = \matr{A}\vect{x}^0 + \vect{b}$, where $\vect{x}^0$ is the initial state, $\matr{A}$ is the system matrix, and $\vect{b}$ is a constant vector.
For a transition happening in $[t_{h}, t_{l}]$, the intersection is enclosed by bounding the constant-flow evolution with an abstraction-error set $\set{E}$:
\begin{equation*}
  \forall t\in [t_{h}, t_{l}]: \vect{x}(t) \in \vect{x}^0 + (\matr{A}\vect{x}^0 + \vect{b})t \oplus \set{E}.
\end{equation*}
The accuracy of the approach highly depends on the duration from the first to the last intersection of the reachable set with the guard set;
$\set{E}$ is large if the time interval $[t_{h}, t_{l}]$ is long.

The dynamics scaling approach is another way to avoid geometric intersection operations,
which introduces a scaling phase before the intersection with the dynamics scaled as $f^{s}(\vect{x},\vect{u}) = g(\vect{x})f(\vect{x},\vect{u})$.
The scaling function $g(\vect{x})$ is defined to flatten the set of states against the guard hyperplane during continuous evolution;
the flattened set $\set{R}^{s}_h$ crosses the guard within a short period $[t_{h}^s, t_{l}^s]$, and thus the intersection can be tightly enclosed by $\set{R}([t_{h}^s, t_{l}^s])$.
Due to the nonlinearity introduced by the scaling function, the scaling phase basically reduces to the reachability analysis for nonlinear dynamics with complexity $\mathcal{O}(n^5)$ \citep[Sec.~5]{althoffReachabilityAnalysisNonlinear2013};
the nonlinear dynamics leads to a large over-approximation in $\set{R}^{s}_h$ if it takes long to flatten the set.

\begin{figure}[t]
  \centering
  \psfragfig[width=0.97\columnwidth]{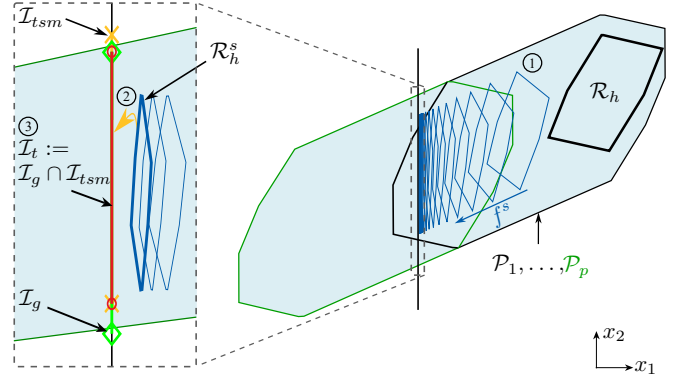}{%
    \psfrag{a}[lc][lc][0.85]{$x_1$}
    \psfrag{b}[lc][lc][0.85]{$x_2$}
    \psfrag{x}[rc][rc][0.85]{$\set{I}_g$}
    \psfrag{y}[rc][rc][0.85]{\shortstack[l]{$\set{I}_t := $ \\ $\set{I}_{g} \cap \set{I}_{tsm}$}}
    \psfrag{z}[cc][cc][0.85]{$\set{R}_{h}$}
    \psfrag{4}[cc][cc][0.85]{}
    \psfrag{5}[rc][rc][0.85]{$\set{I}_{tsm}$}
    \psfrag{6}[lc][lc][0.85]{$\set{R}^{s}_h$}
    \psfrag{7}[lc][lc][0.85]{$\set{P}_1,\ldots,$}
    \psfrag{8}[lc][lc][0.85]{\color[RGB]{0,158,0}{$\set{P}_p$}}
    \psfrag{9}[lc][lc][0.85]{\color[RGB]{0, 92, 171}{$f^{s}$}}
  }
  \caption{Guard intersection with three methods combined.
  }
  \vspace{0em}
  \label{fig_intersect}
\end{figure}

We combine the three approaches to compensate for their individual shortcomings, as shown in Fig.~\ref{fig_intersect}:
We use the scaling approach to flatten the set as the blue sets show, then map the flattened set $\set{R}^{s}_h$ to the guard hyperplane using the mapping method as presented by the yellow arrow. 
The scaling phase shortens the duration of the intersection for the mapping method, while the latter allows terminating the scaling phase before the set expands too much.
We refer to the above combination as time-scaling-mapping (TSM), whose result is denoted by $\set{I}_{tsm}$.
Meanwhile, we compute the geometric intersection $\set{I}_g$
and combine the result to $\set{I}_t := \set{I}_{g} \cap \set{I}_{tsm}$;
the resulting method is referred to as trinal intersection.

Additionally, we introduce two measures to determine when to terminate the scaling phase for TSM.
The first measure evaluates whether the intersection duration is short enough by estimating the time required by a set $\set{X}$ to cross a guard hyperplane \mbox{$\set{G}:=\left\{\vect{x}  \,\middle|\,  \vect{c}^\mathsf{T}\vect{x} = d \right\}, \|\vect{c}\|  = 1$}, given the center of the input set $\vect{u}_c$:
\begin{equation*}
  r^\delta(\set{X}) = \frac{\delta}{\left\lvert c{}^\mathsf{T}  \dot{\vect{x}}_c \right\rvert } = \frac{\mathrm{sup}(\text{\tt{dist}}(\set{X},\set{G}))-\mathrm{inf}(\text{\tt{dist}}(\set{X},\set{G}))}{\left\lvert c{}^\mathsf{T}  f(\text{\tt{center}}(\set{X}),\vect{u}_c) \right\rvert },
\end{equation*}
where $\delta$ represents the length of the set towards the normal direction of the hyperplane while ${\text{\tt{dist}}}(\set{X},\set{G})$ returns the possible distances from states $\vect{x}\in\set{X}$ to hyperplane $\set{G}$ as an interval;
the denominator approximates the traversing speed by projecting the flow of the center point $\dot{\vect{x}}_c$ onto the normal direction of the hyperplane.
The second measure evaluates the expansion of the reachable set to limit the over-approximation in the guard intersection, for which sets are projected onto a plane perpendicular to the flow:
\begin{equation*}
  r^v(\set{X}) = \frac{\text{\tt{vol}}((\matr{I}_n-\dot{\vect{x}}_c {}\dot{\vect{x}}_c {}^\mathsf{T}/\| \dot{\vect{x}}_c \|^2)\otimes\set{X} )}{\text{\tt{vol}}((\matr{I}_n-\dot{\vect{x}}_c {}\dot{\vect{x}}_c {}^\mathsf{T}/\| \dot{\vect{x}}_c \|^2)\otimes\set{R}_h)},
\end{equation*}
where {\text{\tt{vol}}} returns the volume of the projected set, and $\set{R}_h$ is the initial set of the scaling phase.

The implementation of our trinal approach is presented in Alg. \ref{alg2}, involving the following steps (the line numbers coincide with the steps):
\begin{enumerate}
  \item Obtain scaled system dynamics $f^{s}(\vect{x},\vect{u})$,
        with scaling function $g(\vect{x}) = k_s \text{\tt{dist}}(\vect{x},\set{G}) / \mathrm{sup}(\text{\tt{dist}}(\set{R}_h,\set{G}))$,
        where $k_s$ is a gain for tuning.
  \item Obtain a flattened set $\set{R}^{s}_h$ through a scaling phase, 
        where the operator $\text{\tt{reachUntil}}$ computes the reachable sets of the scaled dynamics starting from $\set{R}_h$ 
        until $r^\delta$ decreases to the desired threshold $r^\delta_{d}$ or $r^v$ reaches its limit $r^v_{d}$.
  \item Operation $\text{\tt{refine}}$ computes the reachable sets starting from $\set{R}^{s}_h$ with a finer time step size
        until the set leaves the invariant, in order to refine the list of reachable sets $\set{P}_1^{s},\ldots, \set{P}_q^{s}$ that intersect the guard.
        An approximative intersection duration ${T^s}$ is obtained as the duration between the first step intersecting the guard and the first step after leaving the invariant.
  \item Operation $\text{\tt{input}}$ extracts the input set involved in the intersection, using \eqref{eq_ud} and \eqref{eq_u} where the time interval is $[\mathrm{inf}(\text{\tt{proj}}_t(\set{P}_1^{s})), \mathrm{sup}(\text{\tt{proj}}_t(\set{P}_q^{s}))]$. 
  \item Map $\set{R}^{s}_h$ to the hyperplane using the mapping method (operation \text{\tt{map}}) to obtain the result of the TSM approach $\set{I}_{tsm}$.
\end{enumerate}
Meanwhile, we compute the geometric intersection $\set{I}_g$.
The final result $\set{I}_{tsm} \cap \set{I}_{g}$ can be efficiently computed by enclosing zonotopes by their interval hulls.

\vspace{-0.0em}
\begin{algorithm}[h]
  \caption{\tt{intersectTrinal}}   \label{alg2}
  \begin{algorithmic}[1]
    \renewcommand{\algorithmicrequire}{\textbf{Input:}}
    \renewcommand{\algorithmicensure}{\textbf{Output:}}
    \REQUIRE $f, \set{G}, \lst{P}=(\set{P}_1,\ldots,\set{P}_p), \set{R}_h, \lst{U}_d,\set{U}_u$
    \ENSURE  $\set{I}_t$
    \STATE $f^{s}(\vect{x},\vect{u}) \leftarrow f(\vect{x},\vect{u}) \mkern3mu k_s \text{\tt{dist}}(\vect{x},\set{G}) / \mathrm{sup}(\text{\tt{dist}}(\set{R}_{h},\set{G}))$
    \STATE $\set{R}^{s}_h \leftarrow \text{\tt{reachUntil}}(f^{s}, \set{G}, \set{R}_h,r_{d}^\delta,r_{d}^v)$
    \STATE $(\set{P}_1^{s},\ldots, \set{P}_q^{s},{T^s}) \leftarrow \text{\tt{refine}}(f, \set{G}, \set{R}^{s}_h)$
    \STATE $\set{U} \leftarrow \text{\tt{input}}(\lst{U}_d,\set{U}_u,\mathrm{inf}(\text{\tt{proj}}_t(\set{P}_1^{s})), \mathrm{sup}(\text{\tt{proj}}_t(\set{P}_q^{s})))$
    \STATE $\set{I}_{tsm} \leftarrow \text{\tt{map}}(\set{R}^{s}_h, \set{R}^{s}_{h,l}, f, \set{U}, {T^s})$
    \STATE $\set{I}_t \leftarrow  \set{I}_{tsm} \cap \text{\tt{intersectGeometric}}(\lst{P},\set{G})$
    \RETURN $\set{I}_t$
  \end{algorithmic}
\end{algorithm}

\section{Numerical Examples} \label{sec_example}
We demonstrate the usefulness of our approach by verifying collision safety for HRI \citep{kirschnerExperimentalAnalysisImpact2021}.
We first define the system parameters for the model presented in Sec.~\ref{sec_system}, 
then present the computation results in comparison with the relevant state-of-the-art approaches.
All computations were done in MATLAB based on the toolbox CORA \citep{althoffIntroductionCORA20152015}, on a \SI[]{2.5}[]{GHz} i9 processor and \SI[]{32}[]{GB} memory.
\begin{figure*}[h]
  \centering
  \psfragfig[width=2.0\columnwidth]{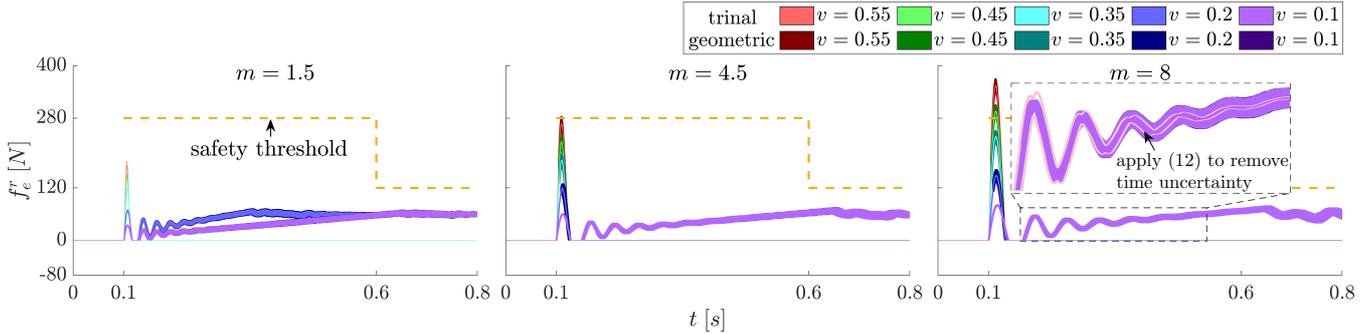}{%
    \psfrag{a}[lt][lt][0.7]{\shortstack[l]{apply \eqref{eq_timehyperplane} to remove \\ time uncertainty}}
  }
  \vspace{-0.5em}
  \caption{Reachability of the contact force with various effective masses and collision speeds.
    Results are shown for both the trinal and the geometric approaches.
  }
  \vspace{0em}
  \label{fig_ra_massspeed}
\end{figure*}
\subsection{System Parameters} \label{sec_sp}
The parameters of the system in Sec.~\ref{sec_system} are listed in Tab.~\ref{table_params}.
Three cases of effective mass are considered, according to the mass distribution of typical robots \citep{kirschnerNotionCorrectUse2021}.
The controller gains in task execution are adopted from the work of DLR \cite[p.~4]{albu-schafferCartesianImpedanceControl2003}, and the damping factor is defined as
$d_t = 2\sqrt{m k_t}$.
For the collision reaction mechanism, the damping factor is tuned by the authors,
and the force threshold of \SI{100}{\newton} is a typical value of robots in the market \citep{kirschnerExperimentalAnalysisImpact2021}.
Regarding the contact dynamics, 
we simplify the sphere to a point mass as its geometry does not affect the system behavior.
The contact coefficients are defined to represent human hands as in ISO/TS 15066 \citep{ISOTS150662016}.
Time delays are quantified based on the controller cycle time $T_c$ and servo cycle time $T_s$ of DLR LWR III \citep{albu-schafferDLRLightweightRobot2007}
with approximations
${d_{rc} \approx 2T_s \approx 0.0006, d_{cr} \approx T_c + T_s \approx 0.0013}$.

\begin{table}[h]
  \vspace{0em}
  \centering
  \caption{System parameters in SI units.}
  \label{table_params}
  \setlength{\tabcolsep}{2.5pt}
  \begin{tabular}{cccccccccccc}
    \toprule
    $m$    & $k_t$ & $d_t$      & $d_r$ & $f_t$ & $k_e$ & $d_e$ & $l$ & $d_1$ & $d_2$ \\
    \midrule
    1.5/4.5/8 & 1000  & 80/135/180 & 380   & 100   & 75000 & 0     & 0   & 0.0013   & 0.0019   \\
    \bottomrule
  \end{tabular}
\end{table}

The discrete-time input $\vect{u}_{d,k}$ is defined as a list of trajectory points sampled at \SI[]{1}[]{kHz},
which hits the surface at \SI[]{0.1}[]{\second} with speed $v$
and stops at $z_d = -0.06$,
and a zonotope $( [0 \mkern5mu 0 \mkern5mu 0]^\mathsf{T}, [0.00005 \mkern5mu 0 \mkern5mu 0]^\mathsf{T})$ following Def. \ref{def_zonotope} describes the input uncertainty $\set{U}_u$ due to the uncertainties in the reference frame.
The initial set of states is represented by a zonotope $( \vect{u}_{d,0}, \text{\tt{diag}}([0.0001 \mkern5mu 0.002 \mkern5mu 0.0001 \mkern5mu 0.002 \mkern5mu 0]^\mathsf{T}))$,
where the operator \text{\tt{diag}} returns a diagonal matrix with the vector elements on the main diagonal.
The time step size $\Delta \tau$ is \SI[]{6.5e-4}[]{\second}, which is derived from half of the delay $d_{cr}$ to avoid large errors in the delayed states.

\subsection{Reachable Set Computation}
We analyze the system for \SI[]{0.8}[]{s} in 15 cases obtained by combining three effective masses (1.5, 4.5, 8 \SI[]{}[]{kg}) and five collision speeds (0.1, 0.2, 0.35, 0.45, 0.55 \SI[]{}[]{m/s}).
The contact force can be easily obtained as described in \eqref{eq_fe} using the linear transformation of zonotopes; the results are shown in Fig.~\ref{fig_ra_massspeed},
which presents very similar behaviors with regard to the experimental results in \cite{kirschnerExperimentalAnalysisImpact2021}:
Both the robot mass and the collision speed have a positive correlation with the contact force reached.
If the threshold of \SI[]{100}[]{N} is not met, the robot tends to stay in contact with the surface, which is the case for low collision speeds such as the purple sets.
When the collision reaction is triggered, the contact force decreases to zero; the robot leaves the surface due to the contact force and stops.
The dotted lines in Fig.~\ref{fig_ra_massspeed} represent the safety limits for human hands \citep{kirschnerExperimentalAnalysisImpact2021}:
\SI[]{280}[]{N} for transient contact (0 to \SI[]{0.5}[]{s} of the contact) and \SI[]{120}[]{N} for quasi-static contact (after \SI[]{0.5}[]{s}),
hence the states with an $f_e^r$ above the limits form an unsafe set $\set{F}$.
By checking whether the reachable sets intersect the thresholds, we can easily verify whether a robot is safe for HRI under a certain speed, and thus define the permissible speed of the robot or fine-tune the controllers.
The verification obviously benefits from the accuracy of guard intersection: $m=4.5, v=0.55$ can be verified as safe by our trinal approach, while the geometric approach is not able to verify its safety.

With case $m=8, v=0.1$ we demonstrate the effect of the technique introduced in Sec. \ref{sec_disctraj}, which removes the time uncertainty after guard intersection.
The boundaries of the resulting reachable sets are plotted in pink in Fig.~\ref{fig_ra_massspeed}.
It can be seen that the additional operation leads to more conservative results for the approximately \SI[]{0.05}[]{s} after the intersection due to a larger initial set, while a better accuracy can be observed in the long term due to a smaller input uncertainty.
In practice, the verification can be more accurate using both results: Since all possible states are enclosed by their intersection, the specifications are violated only if both results intersect the unsafe set.

\begin{table}[h]
  \vspace{0em}
  \setlength{\tabcolsep}{3.3pt}
  \renewcommand{\arraystretch}{0.7}
  \tiny
  \centering
  \setlength{\extrarowheight}{0pt}
  \addtolength{\extrarowheight}{\aboverulesep}
  \addtolength{\extrarowheight}{\belowrulesep}
  \setlength{\aboverulesep}{2pt}
  \setlength{\belowrulesep}{0pt}
  \caption{The volume of guard intersections.}
  \label{table_comparison_vol}
  \begin{tabular}{crclrclrclrclrcl}
    \toprule
                                                                   & \multicolumn{15}{c}{\textbf{\textbf{set volume of the second / third intersection $\times\mathbf{10^3}$}}}                                                                                                                                                                                                                                                                                                                                                                                                                                                                     \\
    \cmidrule(l){2-16}
    \textbf{$m$ \hspace{-0.8pt}/ \hspace{-0.8pt}$v$}\hspace{6.5pt} & \multicolumn{3}{c}{geometric}                                                                              & \multicolumn{3}{c}{mapping}           & \multicolumn{3}{c}{scaling} & \multicolumn{3}{c}{TSM}    & \multicolumn{3}{c}{trinal}                                                                                                                                                                                                                                                                                                                                     \\
    \midrule
    1.5 / 0.10                                                     & 2.17 \hspace{-5pt}                                                                                         & \hspace{-5pt} / \hspace{-5pt}         & \hspace{-5pt} 2.92          & \graycel2.05 \hspace{-5pt} & \hspace{-5pt} \graycel/ \hspace{-5pt} & \hspace{-5pt} \graycel20.0 $^\mathrm{a}$ & \graycel2.71 \hspace{-5pt} & \hspace{-5pt} \graycel/ \hspace{-5pt} & \hspace{-5pt} \graycel135  & \graycel2.35 \hspace{-5pt} & \hspace{-5pt} \graycel/ \hspace{-5pt} & \hspace{-5pt} \graycel16.1 & 1.82 \hspace{-5pt} & \hspace{-5pt} / \hspace{-5pt} & \hspace{-5pt} 2.80 \\
    1.5 / 0.20                                                     & 2.48 \hspace{-5pt}                                                                                         & \hspace{-5pt} / \hspace{-5pt}         & \hspace{-5pt} 3.22          & 1.38 \hspace{-5pt}         & \hspace{-5pt} / \hspace{-5pt}         & \hspace{-5pt} 4.66                       & 1.40 \hspace{-5pt}         & \hspace{-5pt} / \hspace{-5pt}         & \hspace{-5pt} 6.77         & 1.31 \hspace{-5pt}         & \hspace{-5pt} / \hspace{-5pt}         & \hspace{-5pt} 4.76         & 1.21 \hspace{-5pt} & \hspace{-5pt} / \hspace{-5pt} & \hspace{-5pt} 2.51 \\
    1.5 / 0.35                                                     & \graycel7.77 \hspace{-5pt}                                                                                 & \hspace{-5pt} \graycel/ \hspace{-5pt} & \hspace{-5pt} \graycel5.92  & 0.85 \hspace{-5pt}         & \hspace{-5pt} / \hspace{-5pt}         & \hspace{-5pt} 0.21                       & 1.12 \hspace{-5pt}         & \hspace{-5pt} / \hspace{-5pt}         & \hspace{-5pt} 0.40         & 0.81 \hspace{-5pt}         & \hspace{-5pt} / \hspace{-5pt}         & \hspace{-5pt} 0.21         & 0.81 \hspace{-5pt} & \hspace{-5pt} / \hspace{-5pt} & \hspace{-5pt} 0.21 \\
    1.5 / 0.45                                                     & \graycel7.50 \hspace{-5pt}                                                                                 & \hspace{-5pt} \graycel/ \hspace{-5pt} & \hspace{-5pt} \graycel4.42  & 0.57 \hspace{-5pt}         & \hspace{-5pt} / \hspace{-5pt}         & \hspace{-5pt} 0.15                       & 0.58 \hspace{-5pt}         & \hspace{-5pt} / \hspace{-5pt}         & \hspace{-5pt} 0.17         & 0.55 \hspace{-5pt}         & \hspace{-5pt} / \hspace{-5pt}         & \hspace{-5pt} 0.14         & 0.55 \hspace{-5pt} & \hspace{-5pt} / \hspace{-5pt} & \hspace{-5pt} 0.14 \\
    1.5 / 0.55                                                     & \graycel7.63 \hspace{-5pt}                                                                                 & \hspace{-5pt} \graycel/ \hspace{-5pt} & \hspace{-5pt} \graycel4.66  & 0.52 \hspace{-5pt}         & \hspace{-5pt} / \hspace{-5pt}         & \hspace{-5pt} 0.14                       & 0.65 \hspace{-5pt}         & \hspace{-5pt} / \hspace{-5pt}         & \hspace{-5pt} 0.20         & 0.50 \hspace{-5pt}         & \hspace{-5pt} / \hspace{-5pt}         & \hspace{-5pt} 0.13         & 0.50 \hspace{-5pt} & \hspace{-5pt} / \hspace{-5pt} & \hspace{-5pt} 0.13 \\
    4.5 / 0.10                                                     & 1.93 \hspace{-5pt}                                                                                         & \hspace{-5pt} / \hspace{-5pt}         & \hspace{-5pt} 2.57          & \graycel1.90 \hspace{-5pt} & \hspace{-5pt} \graycel/ \hspace{-5pt} & \hspace{-5pt} \graycel22.0               & \graycel2.15 \hspace{-5pt} & \hspace{-5pt} \graycel/ \hspace{-5pt} & \hspace{-5pt} \graycel83.6 & 2.01 \hspace{-5pt}         & \hspace{-5pt} / \hspace{-5pt}         & \hspace{-5pt} 11.2         & 1.52 \hspace{-5pt} & \hspace{-5pt} / \hspace{-5pt} & \hspace{-5pt} 2.48 \\
    4.5 / 0.20                                                     & 7.65 \hspace{-5pt}                                                                                         & \hspace{-5pt} / \hspace{-5pt}         & \hspace{-5pt} 5.44          & 1.06 \hspace{-5pt}         & \hspace{-5pt} / \hspace{-5pt}         & \hspace{-5pt} 0.40                       & 1.17 \hspace{-5pt}         & \hspace{-5pt} / \hspace{-5pt}         & \hspace{-5pt} 0.64         & 1.03 \hspace{-5pt}         & \hspace{-5pt} / \hspace{-5pt}         & \hspace{-5pt} 0.49         & 0.95 \hspace{-5pt} & \hspace{-5pt} / \hspace{-5pt} & \hspace{-5pt} 0.41 \\
    4.5 / 0.35                                                     & 6.09 \hspace{-5pt}                                                                                         & \hspace{-5pt} / \hspace{-5pt}         & \hspace{-5pt} 5.21          & 0.52 \hspace{-5pt}         & \hspace{-5pt} / \hspace{-5pt}         & \hspace{-5pt} 0.24                       & 0.60 \hspace{-5pt}         & \hspace{-5pt} / \hspace{-5pt}         & \hspace{-5pt} 0.31         & 0.49 \hspace{-5pt}         & \hspace{-5pt} / \hspace{-5pt}         & \hspace{-5pt} 0.23         & 0.49 \hspace{-5pt} & \hspace{-5pt} / \hspace{-5pt} & \hspace{-5pt} 0.23 \\
    4.5 / 0.45                                                     & 6.22 \hspace{-5pt}                                                                                         & \hspace{-5pt} / \hspace{-5pt}         & \hspace{-5pt} 5.46          & 0.42 \hspace{-5pt}         & \hspace{-5pt} / \hspace{-5pt}         & \hspace{-5pt} 0.20                       & 0.48 \hspace{-5pt}         & \hspace{-5pt} / \hspace{-5pt}         & \hspace{-5pt} 0.25         & 0.40 \hspace{-5pt}         & \hspace{-5pt} / \hspace{-5pt}         & \hspace{-5pt} 0.20         & 0.40 \hspace{-5pt} & \hspace{-5pt} / \hspace{-5pt} & \hspace{-5pt} 0.20 \\
    4.5 / 0.55                                                     & 6.46 \hspace{-5pt}                                                                                         & \hspace{-5pt} / \hspace{-5pt}         & \hspace{-5pt} 5.80          & 0.39 \hspace{-5pt}         & \hspace{-5pt} / \hspace{-5pt}         & \hspace{-5pt} 0.19                       & 0.42 \hspace{-5pt}         & \hspace{-5pt} / \hspace{-5pt}         & \hspace{-5pt} 0.21         & 0.40 \hspace{-5pt}         & \hspace{-5pt} / \hspace{-5pt}         & \hspace{-5pt} 0.19         & 0.40 \hspace{-5pt} & \hspace{-5pt} / \hspace{-5pt} & \hspace{-5pt} 0.19 \\
    8.0 / 0.10                                                     & 1.82 \hspace{-5pt}                                                                                         & \hspace{-5pt} / \hspace{-5pt}         & \hspace{-5pt} 2.53          & \graycel1.87 \hspace{-5pt} & \hspace{-5pt} \graycel/ \hspace{-5pt} & \hspace{-5pt} \graycel28.9               & \graycel3.11 \hspace{-5pt} & \hspace{-5pt} \graycel/ \hspace{-5pt} & \hspace{-5pt} \graycel31.3 & 2.45 \hspace{-5pt}         & \hspace{-5pt} / \hspace{-5pt}         & \hspace{-5pt} 8.84         & 1.79 \hspace{-5pt} & \hspace{-5pt} / \hspace{-5pt} & \hspace{-5pt} 2.49 \\
    8.0 / 0.20                                                     & 5.87 \hspace{-5pt}                                                                                         & \hspace{-5pt} / \hspace{-5pt}         & \hspace{-5pt} 5.13          & 0.69 \hspace{-5pt}         & \hspace{-5pt} / \hspace{-5pt}         & \hspace{-5pt} 0.34                       & 0.78 \hspace{-5pt}         & \hspace{-5pt} / \hspace{-5pt}         & \hspace{-5pt} 0.47         & 0.56 \hspace{-5pt}         & \hspace{-5pt} / \hspace{-5pt}         & \hspace{-5pt} 0.28         & 0.56 \hspace{-5pt} & \hspace{-5pt} / \hspace{-5pt} & \hspace{-5pt} 0.28 \\
    8.0 / 0.35                                                     & 5.46 \hspace{-5pt}                                                                                         & \hspace{-5pt} / \hspace{-5pt}         & \hspace{-5pt} 5.00          & 0.44 \hspace{-5pt}         & \hspace{-5pt} / \hspace{-5pt}         & \hspace{-5pt} 0.24                       & 0.48 \hspace{-5pt}         & \hspace{-5pt} / \hspace{-5pt}         & \hspace{-5pt} 0.29         & 0.41 \hspace{-5pt}         & \hspace{-5pt} / \hspace{-5pt}         & \hspace{-5pt} 0.22         & 0.41 \hspace{-5pt} & \hspace{-5pt} / \hspace{-5pt} & \hspace{-5pt} 0.22 \\
    8.0 / 0.45                                                     & 5.55 \hspace{-5pt}                                                                                         & \hspace{-5pt} / \hspace{-5pt}         & \hspace{-5pt} 5.20          & 0.38 \hspace{-5pt}         & \hspace{-5pt} / \hspace{-5pt}         & \hspace{-5pt} 0.21                       & 0.62 \hspace{-5pt}         & \hspace{-5pt} / \hspace{-5pt}         & \hspace{-5pt} 0.40         & 0.38 \hspace{-5pt}         & \hspace{-5pt} / \hspace{-5pt}         & \hspace{-5pt} 0.21         & 0.38 \hspace{-5pt} & \hspace{-5pt} / \hspace{-5pt} & \hspace{-5pt} 0.21 \\
    8.0 / 0.55                                                     & 5.88 \hspace{-5pt}                                                                                         & \hspace{-5pt} / \hspace{-5pt}         & \hspace{-5pt} 5.62          & 0.35 \hspace{-5pt}         & \hspace{-5pt} / \hspace{-5pt}         & \hspace{-5pt} 0.20                       & 0.39 \hspace{-5pt}         & \hspace{-5pt} / \hspace{-5pt}         & \hspace{-5pt} 0.23         & 0.35 \hspace{-5pt}         & \hspace{-5pt} / \hspace{-5pt}         & \hspace{-5pt} 0.19         & 0.35 \hspace{-5pt} & \hspace{-5pt} / \hspace{-5pt} & \hspace{-5pt} 0.19 \\
    \bottomrule
    \multicolumn{16}{l}{$^\mathrm{a}$ Failed runs are painted in gray.}
  \end{tabular}
  \bigskip
  \caption{Computation times in seconds.}
  \label{table_comparison_time}
  \small
  \begin{tabular}{cccccc}
    \toprule
                                       & geometric & mapping & scaling & TSM  & trinal \\
    \midrule
    2\textsuperscript{nd} intersection & 0.19      & 0.16    & 13.3    & 1.24 & 1.47   \\
    3\textsuperscript{nd} intersection & 0.31      & 0.17    & 9.81    & 0.55 & 0.77   \\
    \bottomrule                                                                        
  \end{tabular}
\end{table}

We evaluate the accuracy of the proposed guard intersection approaches in comparison with the three methods illustrated in Sec.~\ref{gi}: mapping, scaling, and geometric intersection.
As shown in Tab.~\ref{table_comparison_vol},
TSM shows a clear improvement compared to mapping and scaling.
It also shows overall better accuracy compared to the geometric approach.
Using the latter, the set volume grows significantly during the transition from $\tuple{L}_2$ to $\tuple{L}_3$ due to the over-approximation in the reachable sets of time intervals;
as a result, the geometric approach fails with $m=1.5, v=0.35,0.45,0.55$ where the contact force leads to high acceleration $\ddot{z}^r$.
However, TSM is inaccurate with low-speed collisions ($v=0.1$), where the algorithm cannot effectively shorten the intersection duration, and manual fine-tuning of the scaling function may help. 
The proposed trinal intersection accomplishes all cases
and shows the best accuracy among the tested methods.
It can be seen that, for the trinal approach, TSM determines the final result with good accuracy in most cases,
while the intersection with $\set{I}_{g}$ provides a usable result
even if the intersection duration cannot be effectively shortened before the reachable set overexpands. 

The computation times of guard intersections are shown in Tab.~\ref{table_comparison_time}.
For better readability, we average the time consumed in the runs where all methods pass.
The proposed approaches are significantly more efficient than the scaling approach
because the mapping method and the measures introduced in Sec.~\ref{gi} realize a shorter scaling phase, and the accuracy is even improved due to a tighter flattened set.
The complexity of the two proposed approaches is determined by the mapping method with $\mathcal{O}(n^6)$, which indicates good scalability.

\section{Conclusions}
Results show that our proposed approaches can effectively verify robot behaviors in typical industrial contact tasks.
Contrary to prior work, we overcome numerous challenges associated with real implementations: hybrid dynamics, frequent transitions, strong mechanical impacts, user-specified discrete-time trajectories, and time delays.
Novel approaches for reachability analysis and system modeling are proposed to address these challenges. 
In particular, a novel guard intersection algorithm with complexity $\mathcal{O}(n^6)$ is proposed,
which shows the best performance among the examined state-of-the-art approaches on a fundamental verification case of HRI.
We believe that our work enables formally verifying robot behaviors against contact-task specifications in the real world.
Hence, a subsequent goal is to examine the methods on an abstracted model that formally represents the properties of a real system.
To this end, we will validate the model using conformance checking \citep{roehmReachsetConformanceTesting2016} in future work.

\bibliography{ifacconf}

\end{document}